\documentclass[letterpaper]{article}
\usepackage{aaai2026}
\usepackage{times}
\usepackage{helvet}
\usepackage{courier}
\usepackage[hyphens]{url}
\usepackage{graphicx}
\usepackage{natbib}
\usepackage{caption}
\usepackage{amsmath,amssymb,amsfonts}
\usepackage{algorithm}
\usepackage{algorithmic}
\usepackage{multirow}
\usepackage{booktabs}
\usepackage{textcomp}
\usepackage[table]{xcolor}
\usepackage{pifont}
\usepackage{array}

\newif\ifMMFGUincludeappendix
\MMFGUincludeappendixfalse

\def\BibTeX{{\rm B\kern-.05em{\sc i\kern-.025em b}\kern-.08em
    T\kern-.1667em\lower.7ex\hbox{E}\kern-.125emX}}

\begin{document}

\title{MMFGU: Multimodal Federated Graph Unlearning}

\author{
Haodong Lu, Zekai Chen, Weiwei Ji, Shihao Li, Xunkai Li, Xun Wu, Yinlin Zhu, Rong-Hua Li
}
\affiliations{Beijing Institute of Technology\\
Beijing, China}

\maketitle

\begin{abstract}
Multimodal federated graph learning enables clients to collaboratively train graph models over structural, textual, and visual signals without sharing private local data. 
However, the presence of heterogeneous multimodal content also makes unlearning requests more frequent and fine-grained: users may delete accounts or interactions, remove a particular image or text while retaining the associated entity, or revoke the learned correspondence between retained modalities or graph attributes. 
Existing federated graph unlearning mainly handles entity/relation or client removal and cannot directly satisfy these multimodal requests. 
They introduce three challenges: removing only the requested information without damaging retained content, preventing the target from being recovered through remaining modalities or graph neighborhoods, and stopping related traces on other clients from re-entering the global model after aggregation. 
To address them, we propose \textsc{\textbf{MMFGU}}, a multimodal federated graph unlearning framework built around target-specific representation decoupling. 
\textsc{MMFGU} maps heterogeneous requests into unified target carriers, decouples requested representations while anchoring retained semantics, exposes and repairs propagated residuals with lightweight probes, and selectively purges affected clients through compact prototype and response signals. 
Experiments show that \textsc{MMFGU} effectively removes requested information, preserves retained graph utility, and achieves a $\boldsymbol{41.5\times}$ speedup over full retraining.
\end{abstract}

\begin{figure*}[t]
\centering
\includegraphics[width=0.95\textwidth]{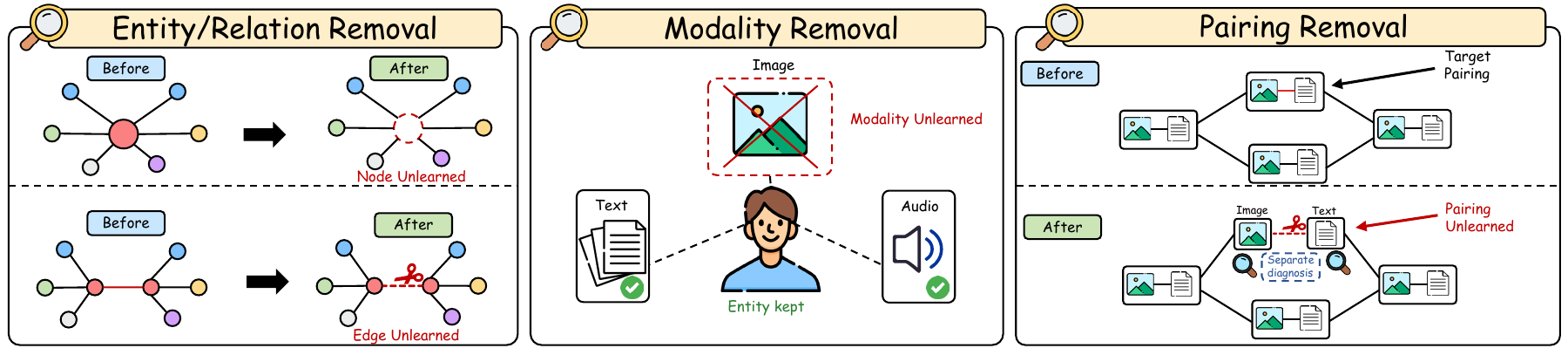}
\caption{Three MM-FGU request types. Entity/Relation Removal deletes a target node or interaction; Modality Removal deletes a specified image, text, or structural-descriptor channel while retaining the entity and its other contents; Pairing Removal revokes a target cross-modal or graph-object--attribute association while retaining both endpoints.}
\label{fig:mmfgu_unlearning_types}
\end{figure*}

\section{Introduction}

Multimodal attributed graphs (MAGs) are widely used to model entities with relational structures and heterogeneous contents, such as social users~\cite{wang2018deep}, citation papers~\cite{yang2016revisiting}, recommendation items~\cite{cai2023lightgcl}, and biomedical entities~\cite{li2023biomedicine}. In practice, these graphs are distributed across platforms, institutions, or devices and may contain sensitive behaviors, private attributes, or proprietary interactions. This motivates multimodal federated graph learning (MM-FGL), which collaboratively trains graph models over structural, textual, and visual signals while keeping raw local graphs private~\cite{openmag,mmopenfgl}. However, multimodal content also makes data-removal requests more frequent and fine-grained: users may delete accounts or interactions, remove an image or profile description while retaining the entity, revoke a specific image--text correspondence, or leave the federation. A deployed MM-FGL model should remove such information without centralizing client data or damaging retained utility. This raises a practical question: how can a multimodal federated graph model selectively forget heterogeneous user requests?

Existing federated graph unlearning (FGU) provides a useful starting point by considering two request scopes: \ding{182} \textbf{Meta Unlearning}, which removes nodes, edges, or features from a client's local subgraph, and \ding{183} \textbf{Client Unlearning}, which erases an entire client's contribution from the global model~\cite{page}. However, such scope-level definitions do not fully specify what must be forgotten in a multimodal graph. We group practical MM-FGU requests into three concrete families, as illustrated in Fig.~\ref{fig:mmfgu_unlearning_types}: \ding{172} \textbf{Entity/Relation Removal}, which deletes a node or interaction; \ding{173} \textbf{Modality Removal}, which deletes one image, text, or structural-descriptor channel while retaining the entity and its other contents; and \ding{174} \textbf{Pairing Removal}, which revokes a specific association, such as an image--text pair or a graph-object--attribute link, while retaining both endpoints. Client removal jointly applies all three removal types to the departing client's contribution.

Although these requests differ in form, they share three challenges. First, \emph{the target influence is distributed}: requested information may already be encoded in modality representations, fusion relations, incident edges, and neighboring nodes. Second, \emph{target and retained representations are entangled}: removing the target should not damage retained modalities, valid correspondences, or unaffected neighborhoods. Third, \emph{residual influence may return through federation}: related visual patterns, textual semantics, relations, or correspondences may remain on other clients and re-enter the global model after aggregation. MM-FGU therefore requires request-aware localization, selective representation decoupling, and cross-client coordination without sharing raw client data.

To address these common challenges, we propose \textsc{MMFGU}, a multimodal federated graph unlearning framework based on target-specific representation decoupling. Rather than designing an independent unlearning rule for each request type, \textsc{MMFGU} maps heterogeneous requests into a unified carrier representation that identifies the affected modality, relation, and neighborhood states. It then moves target carriers toward plausible erased states, while frozen-model anchors preserve retained modalities, correspondences, and graph boundaries. Lightweight perturbation probes further expose and repair target traces that remain in local fusion and neighborhood representations. Finally, prototype-guided screening identifies clients likely to contain target-correlated representations and transfers compact unlearned responses only to these affected clients before selective global aggregation. This pipeline supports diverse MM-FGU requests without sharing raw client data.

\textbf{Our Contributions.}
\underline{\textit{\textbf{(1) Problem Formulation.}}} We formulate MM-FGU with Entity/Relation Removal, Modality Removal, and Pairing Removal requests.
\underline{\textit{\textbf{(2) Framework.}}} We propose \textsc{MMFGU}, combining target decoupling, retention constraints, probe repair, and selective cross-client purge.
\underline{\textit{\textbf{(3) SOTA Performance.}}}
Across \textbf{12 datasets}, \textbf{2 downstream tasks}, and \textbf{3 unlearning request types}, \textsc{MMFGU} achieves the strongest overall utility--unlearning trade-off against \textbf{12 unlearning baselines}, approaching the full-retraining reference in unlearning effectiveness while preserving downstream-task performance and attaining a $\boldsymbol{41.5\times}$ speedup over full retraining.

\section{Related Work}
\begin{figure*}[!t]
    \centering
    \includegraphics[width=\textwidth]{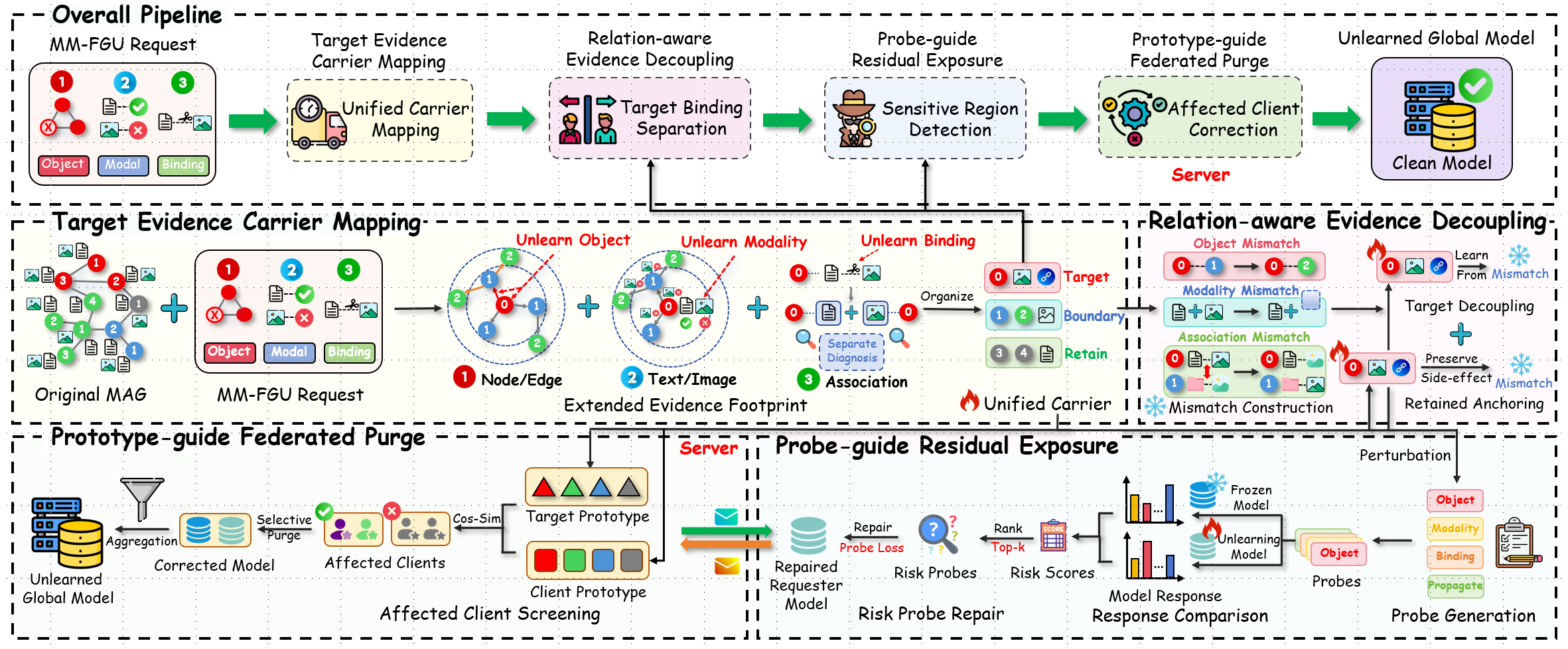}
    \caption{Overall framework of \textsc{MMFGU}. The pipeline maps unlearning requests to target carriers, performs requester-side representation decoupling and probe repair, and coordinates affected clients through prototype-guided federated purge.}
    \label{fig:framework}
\end{figure*}
Existing baselines can be grouped into three categories.
\textbf{General GNN methods} learn node representations through neighborhood aggregation, including convolution-, attention-, and sampling-based architectures~\citep{kipf2017semi,hamilton2017inductive,velickovic2018graph}. Heterophily-aware variants further improve structural modeling under inconsistent neighborhood labels~\citep{pei2020geomgcn,zhu2020gprgnn,chen2020h2gcn}. However, these methods operate on centralized graphs and do not address distributed text-rich attributes.

\textbf{Federated graph learning methods} collaboratively train GNNs over decentralized subgraphs without exchanging raw graph data~\citep{fu2022fgml,fedgraphnn,wu2025comprehensive}. Existing approaches mitigate graph heterogeneity through missing-neighbor reconstruction~\citep{zhang2021subgraph}, personalized modeling~\citep{baek2023personalized}, topology-aware aggregation~\citep{li2024fedgta}, or graph condensation~\citep{fedc4}. Nevertheless, they generally assume compact node features and do not explicitly control the textual evidence retained during graph compression and message passing.

\textbf{Federated TAG methods} introduce language information through local LLM-based augmentation~\citep{yan2025llm4fgl}, globally coordinated codebooks, prompts, or prototypes~\citep{wu2025fedbook,zhu2025fedgfm}, and semantic priors for topology--text interaction. Most of these methods encode text into fixed representations or integrate it into a monolithic graph model, leaving the selection of textual evidence and neighboring nodes largely unconstrained. \textsc{DANCE} instead performs label-aware node condensation together with budgeted, round-adaptive text propagation, enabling efficient federated learning while preserving locally interpretable evidence traces.

\section{Preliminaries}

\subsection{Problem Definition}

\textbf{Federated multimodal graph learning.}
We consider $K$ clients $\mathcal{C}=\{c_1,\ldots,c_K\}$, where client $c_k$ privately holds
\begin{equation}
\mathcal{G}_k=(\mathcal{V}_k,\mathcal{E}_k,
\{\mathbf{X}^{(m)}_k\}_{m\in\mathcal{M}},\mathcal{B}_k), \mathcal{M}=\{s,t,i\}.
\end{equation}
Here, $\mathcal{V}_k$ and $\mathcal{E}_k$ denote local nodes and relations,
$\mathbf{X}^{(m)}_k$ contains structural, textual, or visual features, and
$\mathcal{B}_k$ records modality or graph--attribute bindings.
All raw data remain local. Federated training produces the deployed model
\begin{equation}
\theta^o=\mathcal{A}_{\mathrm{train}}
(\hat\theta;\mathcal{C},\{\mathcal{G}_k\}_{k=1}^{K}).
\end{equation}

\textbf{Unlearning request.}
An MM-FGU request is
\begin{equation}
\begin{aligned}
\mathcal{U}
&=(r,\Delta\mathcal{C},
\{\Delta\mathcal{G}_k\}_{k=1}^{K}),\\
\Delta\mathcal{G}_k
&=(\Delta\mathcal{O}_k,\Delta\mathcal{X}_k,
\Delta\mathcal{B}_k),\\
\Delta\mathcal{O}_k
&=(\Delta\mathcal{V}_k,\Delta\mathcal{E}_k),\\
\Delta\mathcal{X}_k
&\subseteq\mathcal{V}_k\times\mathcal{M}, 
 \Delta\mathcal{B}_k\subseteq\mathcal{B}_k.
\end{aligned}
\end{equation}
where $r$ is the requester and the three deletion components specify
graph objects, modality entries, and bindings.
Meta unlearning has $\Delta\mathcal{C}=\emptyset$ and deletes only items
from $\mathcal{G}_r$, whereas client unlearning sets
$\Delta\mathcal{C}=\{c_r\}$ and removes the full contribution of $c_r$.
The retained federation is
\begin{equation}
\mathcal{C}^{\mathrm{ret}}=\mathcal{C}\setminus\Delta\mathcal{C},
\mathcal{G}_k^{\mathrm{ret}}
=\Pi_{\mathrm{del}}(\mathcal{G}_k;\Delta\mathcal{G}_k).
\end{equation}

\textbf{Unlearning objective.}
Full retraining and practical unlearning respectively produce
\begin{equation}
\begin{aligned}
\theta^\star
&=\mathcal{A}_{\mathrm{train}}
(\hat\theta;\mathcal{C}^{\mathrm{ret}},
\{\mathcal{G}_k^{\mathrm{ret}}\}_{c_k\in\mathcal{C}^{\mathrm{ret}}}),\\
\theta^-
&=\mathcal{A}_{\mathrm{unlearn}}(\theta^o;\mathcal{U}).
\end{aligned}
\end{equation}
Because parameter distance is unreliable for non-convex models, we compare
their behavior on retained and deletion-sensitive queries:
\begin{equation}
\mathcal{D}_{x}(\theta^-,\theta^\star)
=
\mathbb{E}_{q\in\mathcal{Q}_{x}}
d_q\!\left(g_{\theta^-}(q),g_{\theta^\star}(q)\right),
x\in\{\mathrm{ret},\mathrm{del}\}.
\end{equation}
A successful MM-FGU method satisfies
\begin{equation}
\mathcal{D}_{\mathrm{del}}\le\epsilon_{\mathrm{del}},
\mathcal{D}_{\mathrm{ret}}\le\epsilon_{\mathrm{ret}},
\mathrm{Cost}(\mathcal{A}_{\mathrm{unlearn}})
\ll
\mathrm{Cost}(\mathcal{A}_{\mathrm{train}}).
\end{equation}
Unlearning must additionally remove cross-client target-correlated influence
without exposing raw local data.

\subsection{Three Multimodal Unlearning Targets}

MM-FGU considers Entity/Relation Removal, Modality Removal, and Pairing
Removal, corresponding to $\Omega^{\mathsf{obj}}$,
$\Omega^{\mathsf{mod}}$, and $\Omega^{\mathsf{bind}}$.
They remove selected graph objects, modality entries, or associations,
respectively; client unlearning jointly removes all three carrier types
contributed by the departing client.
\section{Methodology}
\label{sec:methodology}

\subsection{Overview}

As shown in Fig.~\ref{fig:framework}, \textsc{MMFGU} maps Entity/Relation Removal, Modality Removal, and Pairing Removal requests to a unified target-carrier representation, enabling one deletion pipeline while preserving unrelated modalities, associations, and neighborhoods. The pipeline first decouples requested carriers and anchors retained ones to the frozen model, then probes perturbed neighborhoods for propagated remnants, and finally coordinates a selective federated purge among clients with similar carrier prototypes to prevent target-correlated representations from re-entering the global model.

\subsection{Request-Specific Target Carriers}

\textbf{Carrier construction.}
We decompose the local model into modality encoders $E_\theta^{(m)}$, relation fusion $F_\theta$, graph propagation $G_\theta$, and a task head. For modality $m$, $\mathbf e_v^{(m)}=E_\theta^{(m)}(\mathbf x_v^{(m)})$; we summarize the subsequent fusion and propagation by $\Psi_\theta$. A carrier $a$ can therefore denote a node-centered representation, an edge relation, one modality channel, or a multimodal binding. For client $c_k$,
\begin{equation}
    \Omega_k
    =
    \Omega_k^{\mathsf{obj}}
    \cup
    \Omega_k^{\mathsf{mod}}
    \cup
    \Omega_k^{\mathsf{bind}},
    \quad
    \mathbf{z}_{\theta}(a)=\Psi_{\theta}(a;\mathcal{G}_k),
\end{equation}
where $\Omega_k^{\mathsf{obj}}$, $\Omega_k^{\mathsf{mod}}$, and $\Omega_k^{\mathsf{bind}}$ collect the carriers for Entity/Relation Removal, Modality Removal, and Pairing Removal, respectively.

\textbf{Request mapping.}
Given request $\mathcal U$, the requester constructs
\begin{equation}
\Omega_r^f=\Gamma(\mathcal U;\mathcal G_r)=
\begin{cases}
\Omega_{r,\mathrm{obj+inc}}, & \Delta\mathcal O_r,\\
\Omega_{r,\mathrm{channel+fusion}}, & \Delta\mathcal X_r,\\
\Omega_{r,\mathrm{paired}}, & \Delta\mathcal B_r.
\end{cases}
\end{equation}
For object deletion, incident carriers capture target information already propagated to adjacent nodes. For modality deletion, the entity and unrequested channels remain outside $\Omega_r^f$. For binding deletion, both unimodal endpoints are retained and only their learned correspondence is targeted. This distinction is what enables selective rather than indiscriminate unlearning.

\subsection{Relation-Aware Target Decoupling}

A target can remain recoverable through image--text similarity, graph-object--attribute association, or neighborhood representations even after its task logit changes. Starting from frozen $\theta^o$, the requester learns $\theta_r^-$. For each $a\in\Omega_r^f$, $T_{\mathrm{mis}}$ constructs $\tilde a$ by replacing an object with a retained-neighborhood alternative, masking/replacing the requested modality, or swapping one endpoint of a binding. The frozen model provides a stable non-target anchor through
\begin{equation}
    p_{\theta}(b|a)
    =
    \frac{\exp(\kappa(\mathbf{z}_{\theta}(a),\mathbf{z}_{\theta^o}(b))/\tau)}
    {\sum_{u\in\mathcal{C}_a}\exp(\kappa(\mathbf{z}_{\theta}(a),\mathbf{z}_{\theta^o}(u))/\tau)},
\end{equation}
where $\mathcal{C}_a=\{\tilde a\}\cup\mathcal{R}_a$ contains the mismatch and retained candidates. Target representations are decoupled by
\begin{equation}
    \mathcal{L}_{\mathrm{dec}}
    =
    -\frac{1}{|\Omega_r^f|}
    \sum_{a\in\Omega_r^f}
    \log p_{\theta_r^-}(\tilde a|a).
\end{equation}
Unlike confidence suppression, this makes the requested carrier indistinguishable from a plausible mismatch in carrier space.

\textbf{Retention constraints.}
Let $\Omega_r^+$ be sampled retained carriers, $\mathcal M^+(a)$ the modality endpoints that must remain usable, and $\mathcal B_r$ the non-target one-hop boundary. We constrain three complementary forms of drift:
\begin{equation}
\begin{gathered}
\mathcal L_{\mathrm{mm}}
=\mathbb E_{a\in\Omega_r^+}
\|\mathbf z_{\theta_r^-}(a)-\mathbf z_{\theta^o}(a)\|_2^2,\\
\mathcal L_{\mathrm{uni}}
=\mathbb E_{\substack{a\in\Omega_r^f\\m\in\mathcal M^+(a)}}
\|\mathbf e_{\theta_r^-}^{(m)}(a)-\mathbf e_{\theta^o}^{(m)}(a)\|_2^2,\\
\mathcal L_{\mathrm{bd}}
=\mathbb E_{v\in\mathcal B_r}
\|\mathbf h_{\theta_r^-}(v)-\mathbf h_{\theta^o}(v)\|_2^2.
\end{gathered}
\end{equation}
The first preserves retained associations, the second protects unrequested modalities, and the third limits graph-propagation drift. The local objective is
\begin{equation}
\mathcal L_{\mathrm{local}}=\mathcal L_{\mathrm{dec}}
+\alpha\mathcal L_{\mathrm{mm}}+\beta\mathcal L_{\mathrm{uni}}
+\lambda\mathcal L_{\mathrm{bd}}.
\end{equation}
Thus, unlearning acts on modality, fusion, and graph representations rather than only moving the final decision boundary.

\subsection{Probe-Guided Residual Exposure}

Local decoupling can leave traces in nearby graph neighborhoods. The requester therefore builds a compact perturbation pool
\begin{equation}
    \mathcal{P}_r
    =
    \{T_{\xi}(\mathcal{G}_r)
    \mid
    \xi\in\Xi(\Omega_r^f),\ \|\xi\|_0\le B_p\},
\end{equation}
where $T_{\xi}$ masks modalities, replaces binding endpoints, removes target relations, or perturbs one-hop neighborhoods. High-risk probes are selected by
\begin{equation}
    \mathcal{P}_r^{\star}
    =
    \operatorname{TopM}_{P\in\mathcal{P}_r}
    D_{\Sigma}\!\left(
    \Phi_{\theta_r^-}(P),
    \Phi_{\theta^o}(P)
    \right),
\end{equation}
where $\Phi_{\theta}$ collects logits, carrier representations, and association scores. Large discrepancy identifies regions where target and retained representations remain entangled. For each selected probe, the requester aligns its response with the masked or mismatched counterpart:
\begin{equation}
    \mathcal{L}_{\mathrm{probe}}
    =
    \frac{1}{|\mathcal{P}_r^{\star}|}
    \sum_{P\in\mathcal{P}_r^{\star}}
    \left\|
    \Phi_{\theta_r^-}(P)
    -
    \mathrm{sg}\big(\Phi_{\theta_r^-}(\widetilde P)\big)
    \right\|_2^2 .
\end{equation}
The requester minimizes $\mathcal L_{\mathrm{req}}=\mathcal L_{\mathrm{local}}+\lambda_p\mathcal L_{\mathrm{probe}}$. Probe repair therefore focuses computation on exposed residuals rather than retraining the full local graph.

\subsection{Prototype-Guided Federated Purge}

Target-related traces may remain on other clients through carriers that are semantically close to the requested target. Let $\mathcal T_\Omega$ denote the carrier groups for Entity/Relation Removal, Modality Removal, and Pairing Removal defined above. During federated training, each client summarizes every non-empty group $\Omega_k^{(\tau)}$:
\begin{equation}
    \mathbf{p}_{k}^{(\tau)}
    =
    \frac{\sum_{a\in\Omega_k^{(\tau)}}w_a\mathbf{z}_{\theta^o}(a)}
    {\sum_{a\in\Omega_k^{(\tau)}}w_a}.
\end{equation}
For the current request, the requester also forms target prototypes from the requested carriers rather than from its full local distribution. Let $\Omega_{r,f}^{(\tau)}=\Omega_r^f\cap\Omega_r^{(\tau)}$ denote the requested carriers of type $\tau$:
\begin{equation}
    \mathbf{g}_{r}^{(\tau)}
    =
    \frac{\sum_{a\in\Omega_{r,f}^{(\tau)}}w_a\mathbf{z}_{\theta^o}(a)}
    {\sum_{a\in\Omega_{r,f}^{(\tau)}}w_a}.
\end{equation}
Only these compact summaries are shared. With $\mathcal I_{r,k}$ denoting the requested carrier groups that are also present on client $k$, the server selects affected clients by
\begin{equation}
    s_{r,k}
    =
    \max_{\tau\in\mathcal I_{r,k}}
    \left[
    \kappa(\mathbf{g}_{r}^{(\tau)},\mathbf{p}_{k}^{(\tau)})
    -
    \tau_p^{(\tau)}
    \right].
\end{equation}
\begin{equation}
    \mathcal{A}_r
    =
    \left\{
    k\ne r\ \middle|\ 
    \begin{array}{l}
    \mathcal I_{r,k}\neq\emptyset,\\
    s_{r,k}>0
    \end{array}
    \right\}.
\end{equation}
The thresholds $\tau_p^{(\tau)}$ can be tied to a single $\tau_p$ in practice. This screening avoids updating unrelated clients whose overall data distribution is similar but whose relevant target carriers are absent. Each affected client receives selected probe responses, not the requester's raw graph or modality data, and optimizes
\begin{equation}
    \mathcal{L}_{\mathrm{purge}}^{(k)}
    =
    \mathcal{L}_{\mathrm{keep}}^{(k)}
    +
    \lambda_{\mathrm{pg}}
    \mathbb{E}_{P\in\mathcal{P}_r^{\star}}
    \left\|
    \Phi_{\theta_k^-}(P)
    -
    \mathrm{sg}\big(\Phi_{\theta_r^-}(P)\big)
    \right\|_2^2 .
\end{equation}
Here, $\mathcal L_{\mathrm{keep}}^{(k)}$ combines the retained task loss and representation consistency on client $k$, while the second term transfers the requester's unlearned response. The server aggregates the requester, affected clients, and unchanged states:
\begin{equation}
    \theta^-
    =
    \operatorname{Agg}
    \left(
    \theta_r^-,
    \{\theta_k^-\}_{k\in\mathcal{A}_r},
    \{\theta_j^o\}_{j\notin\mathcal{A}_r\cup\{r\}}
    \right).
\end{equation}
Selective purge removes cross-client residuals where prototype similarity indicates likely propagation while limiting communication and collateral utility loss.

\textbf{Client-level requests.}
When a client departs, requester-side decoupling is skipped; its saved carrier prototypes define a composite target, and the same selective purge is applied to affected clients.

\raggedbottom
\setlength{\intextsep}{10pt}
\setlength{\textfloatsep}{8pt}
\setlength{\floatsep}{6pt}
\begin{table*}[!t]
    \centering
    \caption{Node- and relation-unlearning performance. Results are reported as retained accuracy/UR for node unlearning ($10\%$ deletion) and Recall@5/UR for relation unlearning ($20\%$ deletion), in percentages. Excluding Retrain, the best results are highlighted in \colorbox[HTML]{DADADA}{\textbf{bold}}, while the second-best are \underline{underlined}.}
    \label{tab:mmfgu_main}
    \vspace{-1pt}
    \scriptsize
    \setlength{\tabcolsep}{1.2pt}
    \renewcommand{\arraystretch}{1.05}
    \resizebox{\textwidth}{!}{
    \begin{tabular}{c|c|ccccccc}
    \specialrule{1.5pt}{1.5pt}{1.5pt}
    \multicolumn{2}{c|}{\textbf{Category}} &
    \multicolumn{7}{c}{\textbf{Node Classification after Node Unlearning: Acc.}$\uparrow$\textbf{ / UR}$\downarrow$\textbf{ (\%)}} \\
    \cmidrule{1-9}
    & \textbf{Method} &
    \textbf{Movies} & \textbf{Grocery} & \textbf{Toys} &
    \textbf{Ele-Fashion} & \textbf{RedditS} & \textbf{Book-nc} & \textbf{Avg.} \\
    \midrule
    \multirow{1}{*}{\textbf{Reference}}
    & \textbf{Retrain}
    & $52.19_{\scriptstyle \pm 1.94}/0.68_{\scriptstyle \pm 0.30}$
    & $82.46_{\scriptstyle \pm 0.48}/0.31_{\scriptstyle \pm 0.71}$
    & $80.32_{\scriptstyle \pm 0.30}/1.58_{\scriptstyle \pm 1.55}$
    & $96.83_{\scriptstyle \pm 0.26}/0.01_{\scriptstyle \pm 0.13}$
    & $96.21_{\scriptstyle \pm 0.47}/0.51_{\scriptstyle \pm 1.54}$
    & $52.13_{\scriptstyle \pm 0.96}/0.42_{\scriptstyle \pm 0.24}$
    & $76.69/0.59$ \\
    \midrule
    \multirow{8}{*}{\textbf{Baselines}}
    & \textbf{FedEraser}
    & $44.78_{\scriptstyle \pm 0.86}/1.46_{\scriptstyle \pm 0.46}$
    & $71.08_{\scriptstyle \pm 0.32}/0.23_{\scriptstyle \pm 1.06}$
    & $72.29_{\scriptstyle \pm 1.17}/2.48_{\scriptstyle \pm 1.36}$
    & $91.55_{\scriptstyle \pm 0.45}/0.65_{\scriptstyle \pm 0.21}$
    & $94.12_{\scriptstyle \pm 0.22}/0.24_{\scriptstyle \pm 2.12}$
    & $45.11_{\scriptstyle \pm 0.70}/0.39_{\scriptstyle \pm 0.11}$
    & $69.82/0.91$ \\
    
    & \textbf{FUSED}
    & $\underline{51.75_{\scriptstyle \pm 0.40}}/12.67_{\scriptstyle \pm 3.39}$
    & $82.37_{\scriptstyle \pm 0.77}/3.95_{\scriptstyle \pm 0.29}$
    & $81.07_{\scriptstyle \pm 0.66}/5.67_{\scriptstyle \pm 1.36}$
    & \cellcolor[HTML]{DADADA}$\textbf{97.28}_{\scriptstyle \pm \textbf{0.37}}/0.55_{\scriptstyle \pm 0.17}$
    & $\underline{96.26_{\scriptstyle \pm 0.50}}/3.94_{\scriptstyle \pm 0.43}$
    & $\underline{53.97_{\scriptstyle \pm 1.27}}/0.54_{\scriptstyle \pm 0.20}$
    & $\underline{77.12}/4.55$ \\
    
    & \textbf{ReGEnUnlearn}
    & $51.31_{\scriptstyle \pm 1.43}/6.20_{\scriptstyle \pm 2.18}$
    & $\underline{82.41_{\scriptstyle \pm 0.35}}/1.68_{\scriptstyle \pm 0.08}$
    & $\underline{81.32_{\scriptstyle \pm 0.49}}/3.29_{\scriptstyle \pm 1.35}$
    & $95.18_{\scriptstyle \pm 0.34}/0.11_{\scriptstyle \pm 0.34}$
    & $95.69_{\scriptstyle \pm 0.60}/1.08_{\scriptstyle \pm 1.53}$
    & $50.24_{\scriptstyle \pm 0.49}/0.45_{\scriptstyle \pm 0.11}$
    & $76.03/2.14$ \\
    
    & \textbf{MoDe}
    & $45.25_{\scriptstyle \pm 2.23}/8.15_{\scriptstyle \pm 2.43}$
    & $81.41_{\scriptstyle \pm 1.00}/1.61_{\scriptstyle \pm 1.07}$
    & $78.05_{\scriptstyle \pm 0.84}/4.27_{\scriptstyle \pm 1.36}$
    & $97.12_{\scriptstyle \pm 0.27}/0.31_{\scriptstyle \pm 0.19}$
    & $96.18_{\scriptstyle \pm 0.39}/1.46_{\scriptstyle \pm 1.25}$
    & $48.59_{\scriptstyle \pm 3.12}/0.44_{\scriptstyle \pm 0.15}$
    & $74.43/2.71$ \\
    
    & \textbf{SIFU}
    & $46.20_{\scriptstyle \pm 1.24}/0.14_{\scriptstyle \pm 0.97}$
    & $78.07_{\scriptstyle \pm 1.55}/0.63_{\scriptstyle \pm 0.62}$
    & $78.13_{\scriptstyle \pm 0.35}/1.73_{\scriptstyle \pm 1.50}$
    & $93.13_{\scriptstyle \pm 0.91}/0.36_{\scriptstyle \pm 0.35}$
    & $94.27_{\scriptstyle \pm 0.68}/0.16_{\scriptstyle \pm 2.01}$
    & $44.27_{\scriptstyle \pm 1.94}/0.38_{\scriptstyle \pm 0.17}$
    & $72.35/0.57$ \\
    
    & \textbf{FedRecover}
    & $48.31_{\scriptstyle \pm 1.52}/13.80_{\scriptstyle \pm 2.37}$
    & $82.07_{\scriptstyle \pm 1.17}/4.14_{\scriptstyle \pm 0.81}$
    & $78.51_{\scriptstyle \pm 0.58}/6.78_{\scriptstyle \pm 0.74}$
    & $96.75_{\scriptstyle \pm 0.30}/0.44_{\scriptstyle \pm 0.20}$
    & $95.84_{\scriptstyle \pm 0.46}/2.29_{\scriptstyle \pm 1.57}$
    & $47.43_{\scriptstyle \pm 1.90}/0.46_{\scriptstyle \pm 0.17}$
    & $74.82/4.65$ \\
    
    & \textbf{FedKD}
    & $51.71_{\scriptstyle \pm 0.59}/12.45_{\scriptstyle \pm 3.31}$
    & $82.15_{\scriptstyle \pm 0.89}/3.46_{\scriptstyle \pm 0.36}$
    & $80.74_{\scriptstyle \pm 0.50}/5.56_{\scriptstyle \pm 1.22}$
    & $97.23_{\scriptstyle \pm 0.40}/0.50_{\scriptstyle \pm 0.17}$
    & $96.11_{\scriptstyle \pm 0.57}/3.53_{\scriptstyle \pm 0.88}$
    & $52.75_{\scriptstyle \pm 2.39}/0.51_{\scriptstyle \pm 0.20}$
    & $76.78/4.34$ \\
    
    & \textbf{FedOSD}
    & $49.85_{\scriptstyle \pm 0.19}/9.49_{\scriptstyle \pm 1.98}$
    & $\underline{82.41_{\scriptstyle \pm 0.86}}/1.96_{\scriptstyle \pm 0.18}$
    & $79.18_{\scriptstyle \pm 0.32}/3.28_{\scriptstyle \pm 1.39}$
    & $96.05_{\scriptstyle \pm 0.45}/0.03_{\scriptstyle \pm 0.43}$
    & $96.04_{\scriptstyle \pm 0.17}/1.85_{\scriptstyle \pm 0.91}$
    & $48.52_{\scriptstyle \pm 4.23}/0.44_{\scriptstyle \pm 0.18}$
    & $75.34/2.84$ \\
    \midrule
    \multirow{1}{*}{\textbf{Ours}}
    & \textbf{MMFGU}
    & \cellcolor[HTML]{DADADA}$\textbf{51.77}_{\scriptstyle \pm \textbf{0.34}}/5.86_{\scriptstyle \pm 2.55}$
    & \cellcolor[HTML]{DADADA}$\textbf{82.64}_{\scriptstyle \pm \textbf{0.28}}/1.46_{\scriptstyle \pm 0.57}$
    & \cellcolor[HTML]{DADADA}$\textbf{81.42}_{\scriptstyle \pm \textbf{0.61}}/2.84_{\scriptstyle \pm 0.99}$
    & $\underline{97.25_{\scriptstyle \pm 0.41}}/0.54_{\scriptstyle \pm 0.17}$
    & \cellcolor[HTML]{DADADA}$\textbf{96.27}_{\scriptstyle \pm \textbf{0.44}}/1.01_{\scriptstyle \pm 0.84}$
    & \cellcolor[HTML]{DADADA}$\textbf{55.40}_{\scriptstyle \pm \textbf{0.93}}/0.34_{\scriptstyle \pm 0.26}$
    & \cellcolor[HTML]{DADADA}$\textbf{77.46}/2.01$ \\
    \specialrule{1.3pt}{2.0pt}{1.0pt}
    \end{tabular}}
    \par\vspace{8pt}
    \setlength{\tabcolsep}{2.3pt}
    \renewcommand{\arraystretch}{1.05}
    \resizebox{\textwidth}{!}{
    \begin{tabular}{c|c|ccccccc}
    \specialrule{1.5pt}{1.5pt}{1.5pt}
    \multicolumn{2}{c|}{\textbf{Category}} &
    \multicolumn{7}{c}{\textbf{Modality Retrieval after Relation Unlearning: Recall@5}$\uparrow$\textbf{ / UR}$\downarrow$\textbf{ (\%)}} \\
    \cmidrule{1-9}
    & \textbf{Method} &
    \textbf{QB} & \textbf{TN} & \textbf{KU} &
    \textbf{Bili-Food} & \textbf{Bili-Dance} & \textbf{Bili-Movie} & \textbf{Avg.} \\
    \midrule
    \multirow{1}{*}{\textbf{Reference}}
    & \textbf{Retrain}
    & $92.62_{\scriptstyle \pm 0.83}/34.39_{\scriptstyle \pm 2.68}$ & $87.07_{\scriptstyle \pm 1.90}/31.37_{\scriptstyle \pm 1.98}$ & $80.46_{\scriptstyle \pm 1.16}/6.83_{\scriptstyle \pm 2.43}$
    & $93.92_{\scriptstyle \pm 2.09}/11.32_{\scriptstyle \pm 27.32}$ & $92.12_{\scriptstyle \pm 3.54}/0.24_{\scriptstyle \pm 14.84}$ & $90.76_{\scriptstyle \pm 1.99}/23.34_{\scriptstyle \pm 20.50}$ & $89.49/17.92$ \\
    \midrule
    \multirow{8}{*}{\textbf{Baselines}}
    & \textbf{FineTune}
    & $88.64_{\scriptstyle \pm 1.86}/6.31_{\scriptstyle \pm 0.54}$ & $87.47_{\scriptstyle \pm 1.86}/7.51_{\scriptstyle \pm 0.50}$ & $85.48_{\scriptstyle \pm 1.11}/12.24_{\scriptstyle \pm 3.08}$
    & $93.52_{\scriptstyle \pm 4.90}/1.99_{\scriptstyle \pm 8.87}$ & $93.76_{\scriptstyle \pm 1.28}/1.36_{\scriptstyle \pm 6.75}$ & $87.69_{\scriptstyle \pm 1.08}/6.71_{\scriptstyle \pm 10.02}$ & $89.43/6.02$ \\
    & \textbf{NegGrad}
    & $91.48_{\scriptstyle \pm 0.48}/4.29_{\scriptstyle \pm 1.08}$ & $86.03_{\scriptstyle \pm 1.37}/2.58_{\scriptstyle \pm 2.72}$ & $85.07_{\scriptstyle \pm 1.51}/4.77_{\scriptstyle \pm 2.89}$
    & $92.65_{\scriptstyle \pm 3.95}/1.37_{\scriptstyle \pm 7.06}$ & $93.44_{\scriptstyle \pm 0.96}/0.68_{\scriptstyle \pm 2.38}$ & $89.12_{\scriptstyle \pm 2.26}/7.61_{\scriptstyle \pm 1.55}$ & $89.63/3.55$ \\
    & \textbf{L-codec}
    & $92.43_{\scriptstyle \pm 0.26}/5.21_{\scriptstyle \pm 1.42}$ & $86.80_{\scriptstyle \pm 1.65}/5.27_{\scriptstyle \pm 1.05}$ & $86.07_{\scriptstyle \pm 1.54}/7.70_{\scriptstyle \pm 3.23}$
    & $92.26_{\scriptstyle \pm 4.40}/1.76_{\scriptstyle \pm 8.36}$ & $88.21_{\scriptstyle \pm 2.75}/0.49_{\scriptstyle \pm 1.11}$ & $90.07_{\scriptstyle \pm 2.26}/8.71_{\scriptstyle \pm 0.93}$ & $89.31/4.86$ \\
    & \textbf{DtD}
    & $92.14_{\scriptstyle \pm 0.39}/5.20_{\scriptstyle \pm 1.15}$ & $89.29_{\scriptstyle \pm 1.35}/6.20_{\scriptstyle \pm 1.76}$ & $85.21_{\scriptstyle \pm 0.99}/4.97_{\scriptstyle \pm 3.01}$
    & $92.20_{\scriptstyle \pm 4.09}/1.06_{\scriptstyle \pm 6.18}$ & $93.10_{\scriptstyle \pm 0.88}/0.33_{\scriptstyle \pm 0.89}$ & $90.29_{\scriptstyle \pm 2.17}/8.26_{\scriptstyle \pm 1.40}$ & $90.37/4.34$ \\
    & \textbf{EASE}
    & $\underline{92.65_{\scriptstyle \pm 0.52}}/5.12_{\scriptstyle \pm 1.02}$ & $88.92_{\scriptstyle \pm 0.87}/6.63_{\scriptstyle \pm 2.07}$ & $\underline{87.69_{\scriptstyle \pm 1.92}}/7.88_{\scriptstyle \pm 1.99}$
    & $94.28_{\scriptstyle \pm 2.58}/1.88_{\scriptstyle \pm 8.33}$ & \cellcolor[HTML]{DADADA}$\textbf{94.42}_{\scriptstyle \pm \textbf{0.96}}/3.53_{\scriptstyle \pm 0.58}$ & $\underline{90.77_{\scriptstyle \pm 1.68}}/7.77_{\scriptstyle \pm 0.84}$ & $\underline{91.46}/5.47$ \\
    & \textbf{UL}
    & $92.02_{\scriptstyle \pm 0.25}/4.47_{\scriptstyle \pm 1.72}$ & $88.59_{\scriptstyle \pm 1.82}/6.21_{\scriptstyle \pm 1.61}$ & $86.67_{\scriptstyle \pm 2.03}/5.35_{\scriptstyle \pm 3.15}$
    & $93.89_{\scriptstyle \pm 2.99}/2.23_{\scriptstyle \pm 7.59}$ & $87.50_{\scriptstyle \pm 2.21}/0.54_{\scriptstyle \pm 2.27}$ & $89.36_{\scriptstyle \pm 2.00}/8.37_{\scriptstyle \pm 0.52}$ & $89.67/4.53$ \\
    & \textbf{ERM-KTP}
    & $91.47_{\scriptstyle \pm 0.62}/4.96_{\scriptstyle \pm 0.38}$ & \cellcolor[HTML]{DADADA}$\textbf{90.00}_{\scriptstyle \pm \textbf{1.57}}/6.52_{\scriptstyle \pm 1.75}$ & $87.55_{\scriptstyle \pm 2.01}/5.34_{\scriptstyle \pm 2.28}$
    & \cellcolor[HTML]{DADADA}$\textbf{94.73}_{\scriptstyle \pm \textbf{2.48}}/2.12_{\scriptstyle \pm 7.49}$ & $90.82_{\scriptstyle \pm 3.38}/2.08_{\scriptstyle \pm 1.04}$ & $89.35_{\scriptstyle \pm 2.25}/6.77_{\scriptstyle \pm 3.23}$ & $90.65/4.63$ \\
    & \textbf{FedCSA}
    & $91.53_{\scriptstyle \pm 1.40}/4.49_{\scriptstyle \pm 0.92}$ & $85.99_{\scriptstyle \pm 1.92}/6.94_{\scriptstyle \pm 1.31}$ & $83.33_{\scriptstyle \pm 1.27}/5.66_{\scriptstyle \pm 3.29}$
    & $92.80_{\scriptstyle \pm 4.15}/1.18_{\scriptstyle \pm 7.53}$ & $93.43_{\scriptstyle \pm 2.36}/1.85_{\scriptstyle \pm 4.55}$ & $88.87_{\scriptstyle \pm 0.73}/8.72_{\scriptstyle \pm 1.34}$ & $89.33/4.81$ \\
    \midrule
    \multirow{1}{*}{\textbf{Ours}}
    & \textbf{MMFGU}
    & \cellcolor[HTML]{DADADA}$\textbf{92.68}_{\scriptstyle \pm \textbf{0.24}}/4.25_{\scriptstyle \pm 0.61}$ & $\underline{89.63_{\scriptstyle \pm 1.89}}/5.91_{\scriptstyle \pm 2.62}$ & \cellcolor[HTML]{DADADA}$\textbf{87.81}_{\scriptstyle \pm \textbf{2.01}}/3.09_{\scriptstyle \pm 1.73}$
    & $\underline{94.34_{\scriptstyle \pm 2.96}}/0.40_{\scriptstyle \pm 7.07}$ & $\underline{94.09_{\scriptstyle \pm 0.87}}/0.72_{\scriptstyle \pm 1.29}$ & \cellcolor[HTML]{DADADA}$\textbf{91.24}_{\scriptstyle \pm \textbf{1.12}}/6.05_{\scriptstyle \pm 2.05}$ & \cellcolor[HTML]{DADADA}$\textbf{91.63}/3.40$ \\
    \specialrule{1.3pt}{2.0pt}{1.0pt}
    \end{tabular}}
    \vspace{-3pt}
\end{table*}

\section{Experiments}
\label{sec:mmfgu_exp}

We organize the evaluation around four high-level questions.
\textbf{Q1:} Does \textsc{MMFGU} achieve a favorable utility--unlearning trade-off across the three MM-FGU request types?
\textbf{Q2:} Does it selectively remove the requested information, and which components make this possible?
\textbf{Q3:} Is it robust across deletion regimes and federated settings?
\textbf{Q4:} Does it satisfy the efficiency requirements of practical deployment?

\subsection{Experimental Setup}
\label{sec:mmfgu_setup}

\textbf{Datasets and Protocol.}
We use multimodal graphs from OpenMAG and MM-OpenFGL~\cite{openmag,mmopenfgl}: Movies, Grocery, Toys~\cite{ni2019justifying}, Ele-Fashion~\cite{hou2024bridging,ni2019justifying}, RedditS~\cite{desai2021redcaps}, and Book-nc~\cite{wan2018item,wan2019fine} for node classification; and QB, TN, KU, Bili-Food, Bili-Dance, and Bili-Movie~\cite{zhang2025ninerec} for relation prediction. Graphs are partitioned by Louvain~\cite{louvain}. Default node/relation deletion ratios are $0.1/0.2$; results are mean$\pm$std over five seeds with identical requests.

\textbf{Baselines and Metrics.}
We compare with full \textit{Retrain}~\cite{machine_unlearning}; federated unlearning methods~\cite{federaser,fused,regenunlearn,mode,fedkd,fedosd}; and relation/multimodal baselines~\cite{golatkar2020,lcodec,dtd,ul,ermktp,multidelete}.
Retained utility is measured by accuracy or Recall@5.
For attack-based evaluation of unlearning effectiveness, we report \emph{Unlearning Residual} (UR):
\begin{equation}
\mathrm{UR}=100\times\left|\mathrm{AUC}_{\mathrm{MIA}}-0.5\right|.
\end{equation}
UR measures the residual membership distinguishability of unlearned samples and is reported in percentage points. Lower is better; $\mathrm{UR}=0$ means the attack is no better than random guessing.
We additionally report UR, unlearned-set accuracy, modality unlearning drop, association deltas, runtime, and communication cost.

\subsection{Overall Performance (Q1)}
\label{sec:mmfgu_q1}

\begin{figure*}[!t]
    \centering
    \begin{minipage}[t]{0.49\textwidth}
        \centering
        \includegraphics[width=0.49\linewidth]{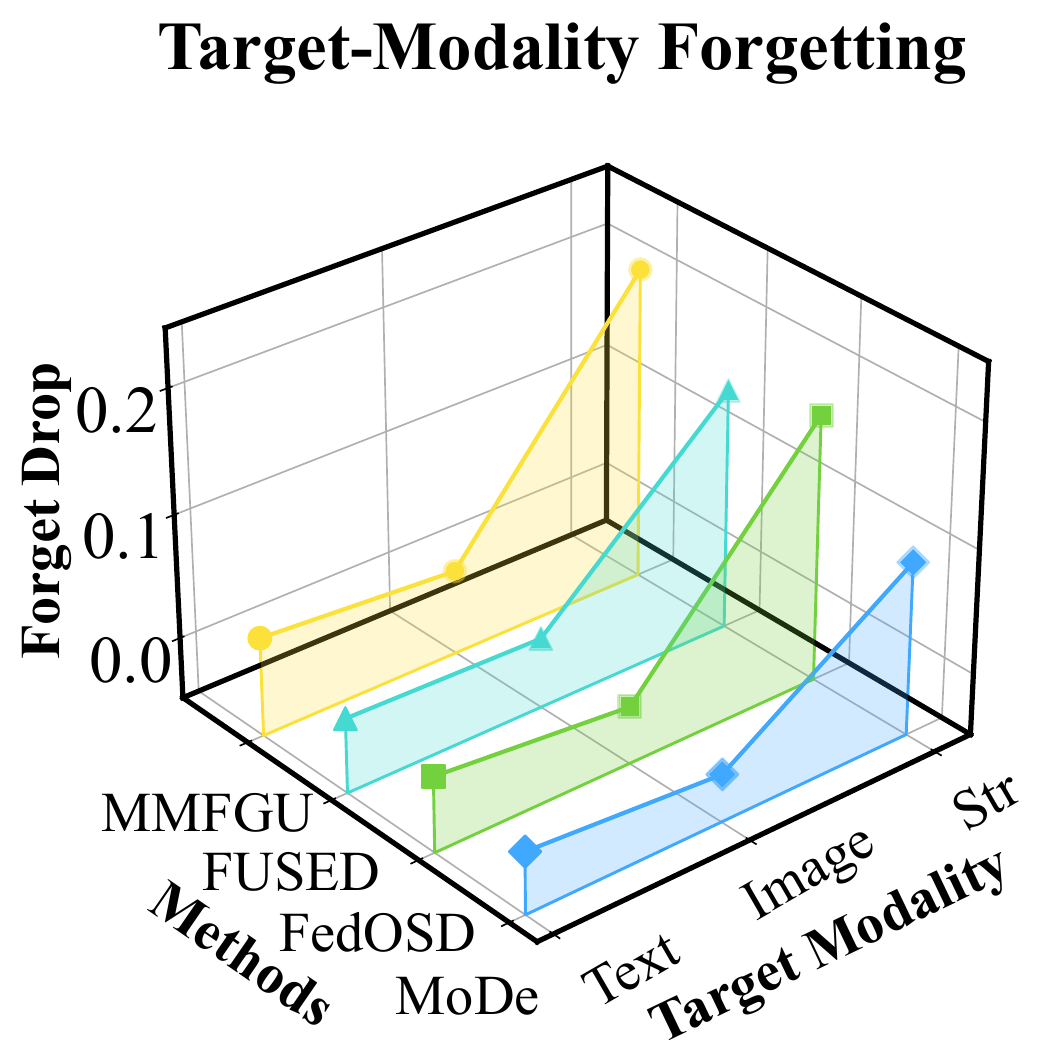}
        \hfill
        \includegraphics[width=0.49\linewidth]{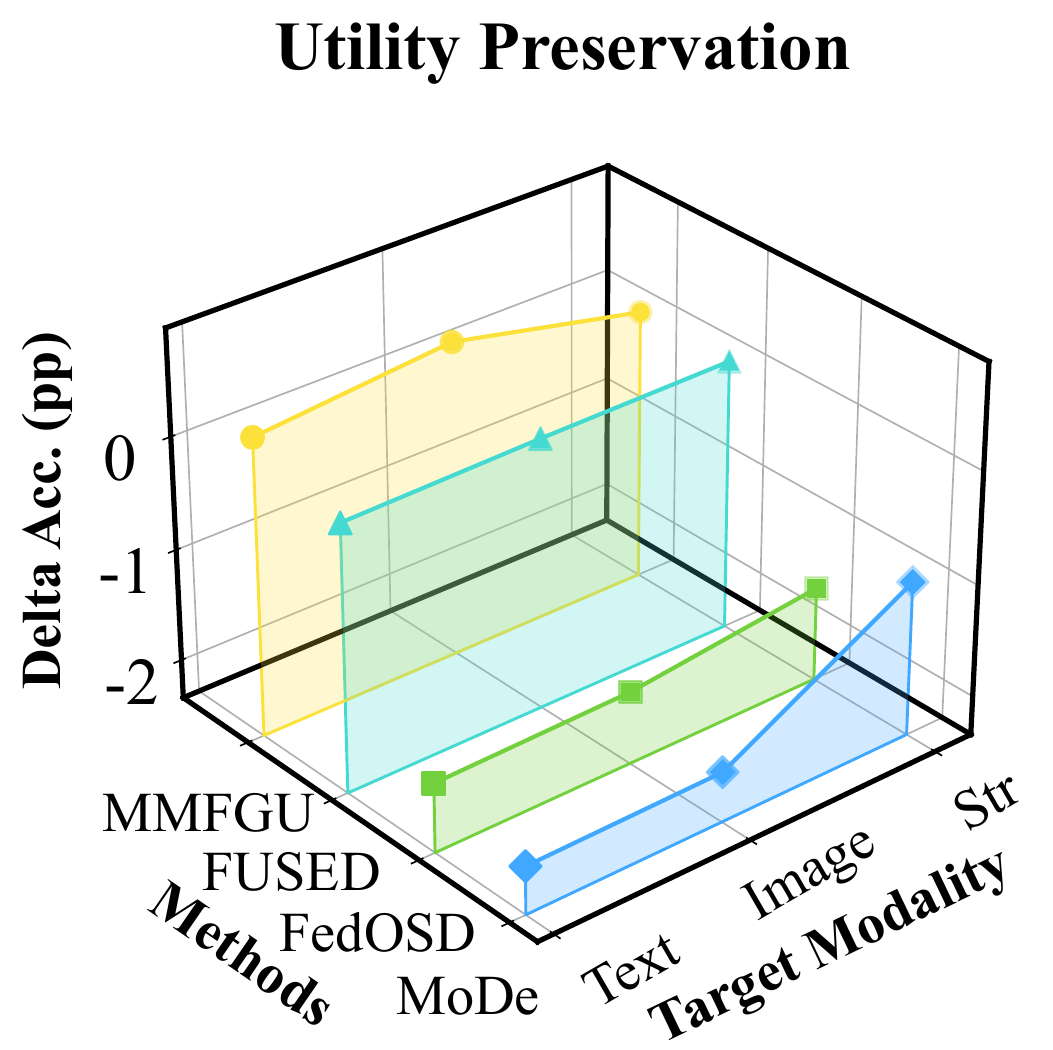}
    \end{minipage}
    \hfill
    \begin{minipage}[t]{0.49\textwidth}
        \centering
        \includegraphics[width=0.49\linewidth]{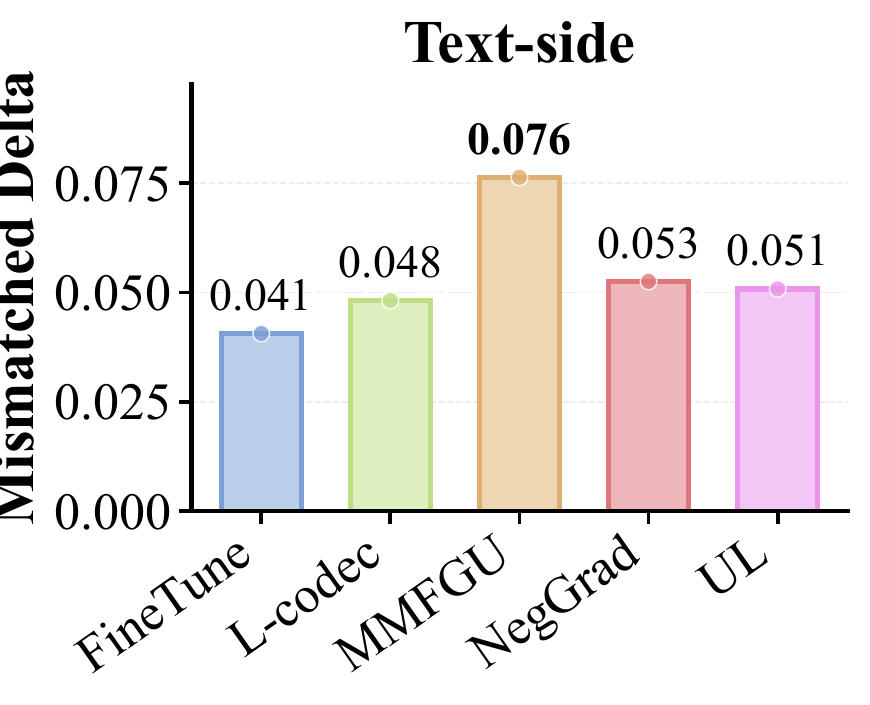}
        \hfill
        \includegraphics[width=0.49\linewidth]{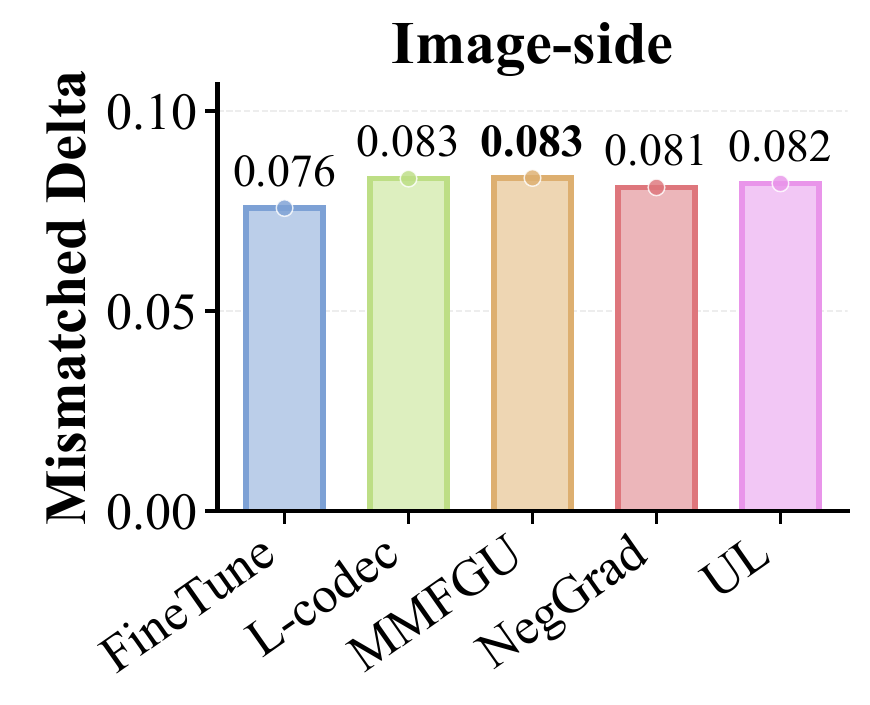}
    \end{minipage}
    \caption{Overall performance on multimodal requests. From left to right: target-modality unlearning drop, retained-utility change, text-side mismatched-association delta, and image-side mismatched-association delta.}
    \label{fig:mmfgu_multimodal_overall}
\end{figure*}

To answer Q1, we compare \textsc{MMFGU} with representative baselines across Entity/Relation Removal, Modality Removal, and Pairing Removal requests, jointly evaluating retained utility and unlearning effectiveness.
Table~\ref{tab:mmfgu_main} provides the primary Entity/Relation Removal comparison, followed by the multimodal results in Fig.~\ref{fig:mmfgu_multimodal_overall}.

\textbf{Entity/Relation Removal Requests.}
In the upper block of Table~\ref{tab:mmfgu_main}, \textsc{MMFGU} achieves the best average retained accuracy ($77.46\%$) across the six node-unlearning datasets.
Its average UR is small in absolute terms but is not the minimum among all node-unlearning baselines; the clearest advantage in this block is therefore retained utility rather than uniform attack-side dominance.
For relation (edge) requests, \textsc{MMFGU} achieves both the best average Recall@5 ($91.63\%$) and the lowest average UR ($3.40\%$), giving the strongest joint utility--unlearning result.

\textbf{Modality Removal and Pairing Removal Requests.}
Fig.~\ref{fig:mmfgu_multimodal_overall} places the two multimodal request types side by side.
Desired outcomes are a large target-modality unlearning drop with near-zero retained-accuracy change for Modality Removal, and a larger mismatched-association delta for Pairing Removal.
\textsc{MMFGU} gives the strongest text-side change and ties the best image-side result.
Overall, the results for Q1 show that \textsc{MMFGU} couples the strongest average retained utility with competitive unlearning on node requests, achieves the best joint relation result, and preserves non-target behavior during Modality Removal and Pairing Removal requests.

\subsection{Unlearning Quality and Mechanism Analysis (Q2)}
\label{sec:mmfgu_q2}

To answer Q2, we examine forget-set behavior, association decoupling, unimodal endpoint retention, and component ablations to determine whether \textsc{MMFGU} removes target information selectively and which modules enable this behavior.

\begin{figure}[!t]
    \centering
    \includegraphics[width=0.94\columnwidth]{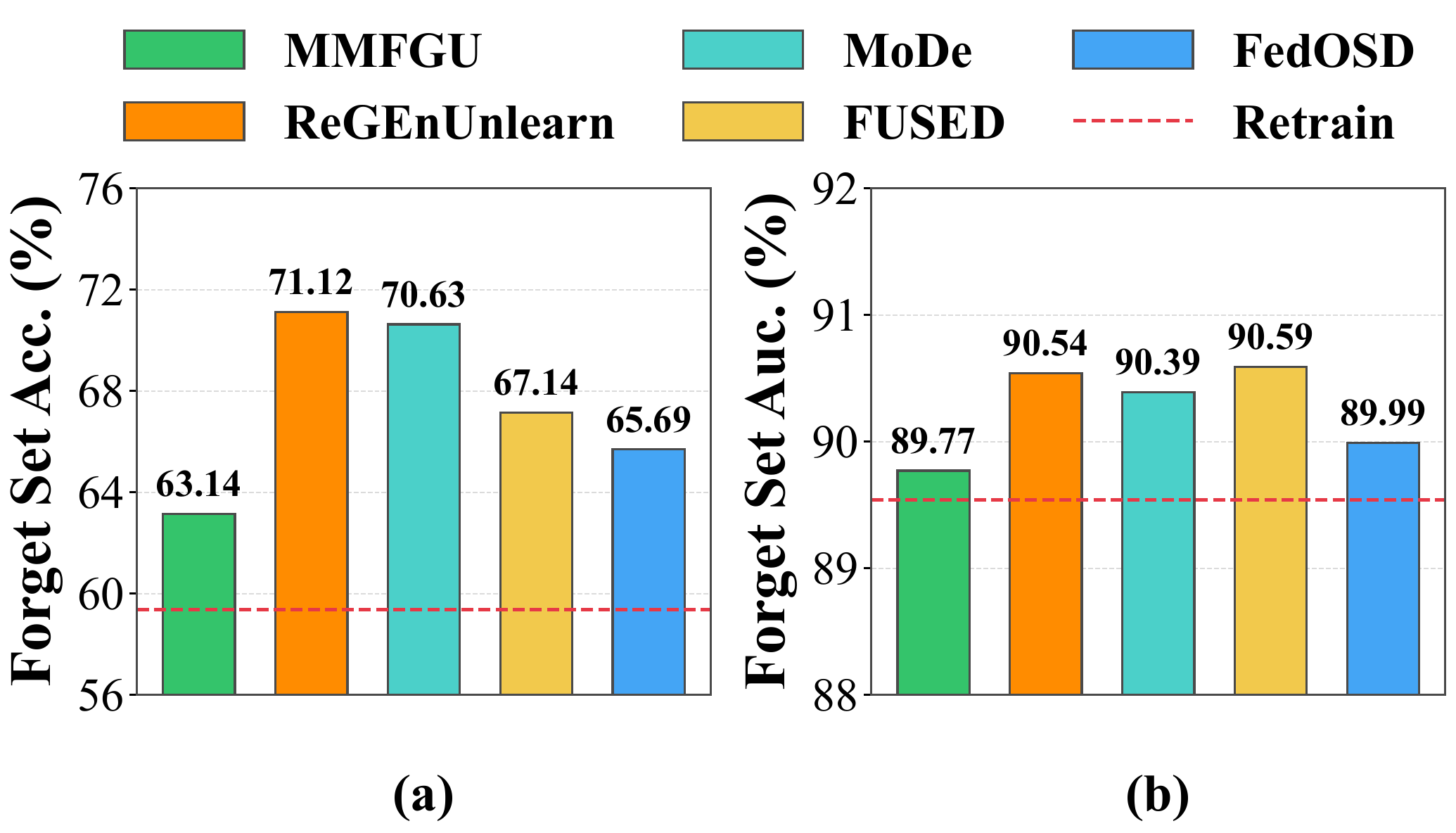}
    \vspace{-10pt}
    \caption{After-forgetting forget-set performance across two tasks: (a) node classification measured by Forget Set Accuracy (\%) and (b) link prediction measured by Forget Set AUC (\%). The red dashed line marks the Retrain baseline.}
    \label{fig:mmfgu_forget_set}
\end{figure}

\textbf{Comparison of Unlearning Effectiveness.}
Fig.~\ref{fig:mmfgu_forget_set} compares the unlearning effect on the deleted nodes: \textsc{MMFGU} approaches Retrain on Movies and gives the lowest unlearned-set accuracy on Grocery.
Together with high retained accuracy, this supports selective target suppression rather than global damage.

\begin{figure}[t]
    \centering
    \includegraphics[width=\columnwidth]{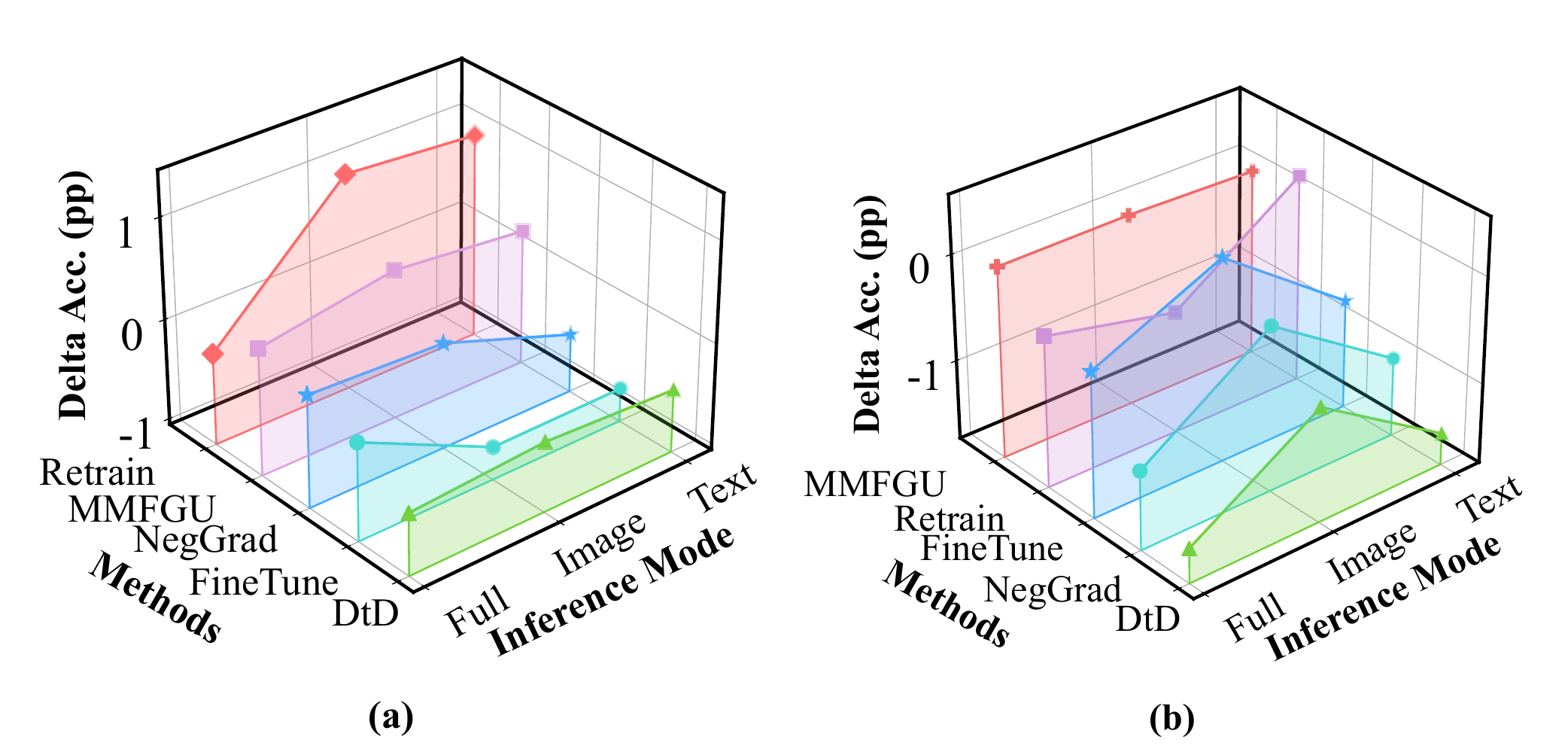}
    \vspace{-20pt}
    \caption{Unimodal endpoint retention on Movies (a) and Grocery (b). Values closer to zero indicate better preservation after Pairing Removal.}
    \label{fig:mmfgu_unimodal_retention}
\end{figure}

\textbf{Association Decoupling and Endpoint Retention.}
We verify whether \textsc{MMFGU} removes target image--text bindings without damaging unimodal endpoints.
Tables~S1--S2 show that unlearned true pairs become closer to random mismatches and that \textsc{MMFGU} achieves stronger decoupling than baselines.
Fig.~\ref{fig:mmfgu_unimodal_retention} shows that full, image-only, and text-only accuracies remain stable, with at most $0.42$ percentage-point change.
Together with Fig.~\ref{fig:mmfgu_multimodal_overall}, these results indicate selective Pairing Removal rather than global representation collapse.

\textbf{Component and Cross-client Analysis.}
Table~\ref{tab:mmfgu_component_ablation} shows that each component contributes to maintaining retained accuracy after unlearning.
Among them, cross-client purge and probe alignment are the most critical: removing either of them leads to a clear and consistent performance drop across datasets, suggesting that requester-local decoupling alone cannot ensure stable global unlearning.
In contrast, removing local decoupling causes a milder but still consistent degradation.
These results indicate complementary roles of the three modules: local objectives initiate target removal, while probe alignment and cross-client propagation prevent residual target information from reappearing after aggregation.
\begin{table}[t]
    \centering
    \caption{Ablation study. Each variant removes one core module to validate its contribution. The best results are highlighted in \colorbox[HTML]{DADADA}{\textbf{bold}}, while the second-best in \underline{underline}.}
    \label{tab:mmfgu_component_ablation}
    \resizebox{\columnwidth}{!}{
    \begin{tabular}{c|cccc}
    \specialrule{1.5pt}{1.5pt}{1.5pt}
    \textbf{Variants}
    & \textbf{Movies}
    & \textbf{Grocery}
    & \textbf{Toys}
    & \textbf{RedditS} \\
    \midrule

    \cellcolor[HTML]{DADADA}\textbf{Full \textsc{MMFGU}}
    & \cellcolor[HTML]{DADADA}\textbf{51.43}
    & \cellcolor[HTML]{DADADA}\textbf{82.80}
    & \cellcolor[HTML]{DADADA}\textbf{81.41}
    & \cellcolor[HTML]{DADADA}\textbf{96.28} \\

    w/o local decouple
    & \underline{50.61}
    & \underline{79.35}
    & \underline{79.83}
    & \underline{96.06} \\

    w/o retain/boundary
    & 48.48
    & 76.75
    & 77.26
    & 93.56 \\

    w/o probe
    & 49.79
    & 77.94
    & 79.03
    & 92.23 \\

    w/o cross-client purge
    & 46.71
    & 77.75
    & 75.49
    & 91.90 \\

    w/o probe alignment
    & 45.25
    & 75.06
    & 77.55
    & 89.36 \\

    \specialrule{1.3pt}{2.0pt}{1.0pt}
    \end{tabular}}
\end{table}

\textbf{Additional Diagnostics.}
Paired-random distance (Table~S1 in the supporting material), decoupling-baseline comparisons (Table~S2 in the supporting material), unimodal retention (Fig.~S1 in the supporting material), and the target-neighbor case study (Fig.~S4 and Table~S5 in the supporting material) provide additional diagnostics.

\subsection{Robustness Analysis (Q3)}
\label{sec:mmfgu_q3}

To answer Q3, we vary unlearning ratios and hyperparameters and integrate \textsc{MMFGU} with different unlearning backbones to assess its robustness across deletion regimes and federated settings.

\begin{figure}[H]
    \centering
    \includegraphics[width=0.49\linewidth]{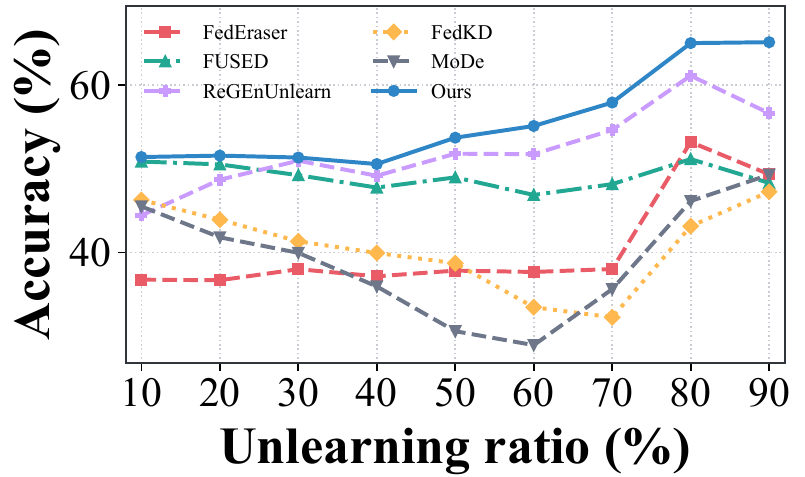}
    \hfill
    \includegraphics[width=0.49\linewidth]{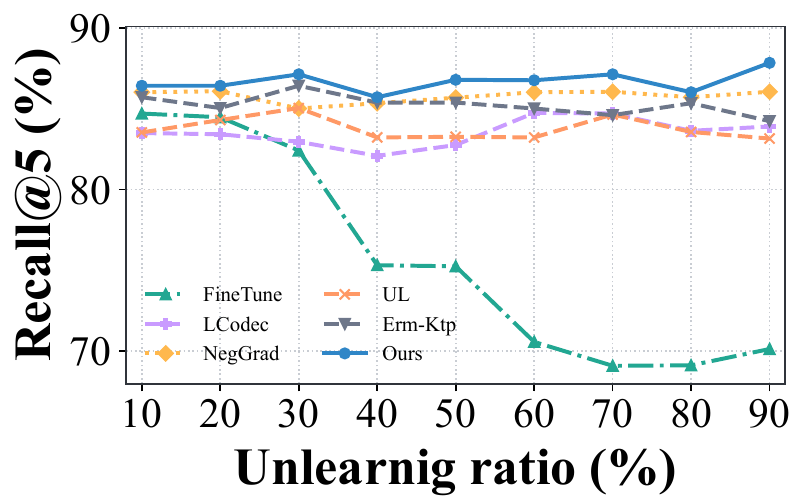}
    \vspace{-8pt}
    \caption{Robustness to the Entity/Relation Removal ratio. Left: Book-nc node classification under different client-unlearning ratios. Right: TN recommendation under different relation-unlearning ratios.}
    \label{fig:mmfgu_ratio_robustness}
\end{figure}

\textbf{Robustness to Unlearning Ratio.}
As shown in Fig.~\ref{fig:mmfgu_ratio_robustness}, \textsc{MMFGU} remains top-ranked across the tested ratios on both Book-nc and TN.
Its Book-nc accuracy changes with the retained-client composition, whereas its TN Recall@5 remains within a narrow band.
Thus, the observed results support a consistent ranking across regimes rather than a uniformly flat absolute curve.
The contrast is especially clear at high unlearning ratios: several baselines deteriorate sharply on at least one task, while \textsc{MMFGU} retains its utility advantage through the $90\%$ endpoint.
This indicates that its behavior is not tied to the default deletion ratio used in Table~\ref{tab:mmfgu_main}.

\begin{figure}[H]
    \centering
    \includegraphics[width=\columnwidth]{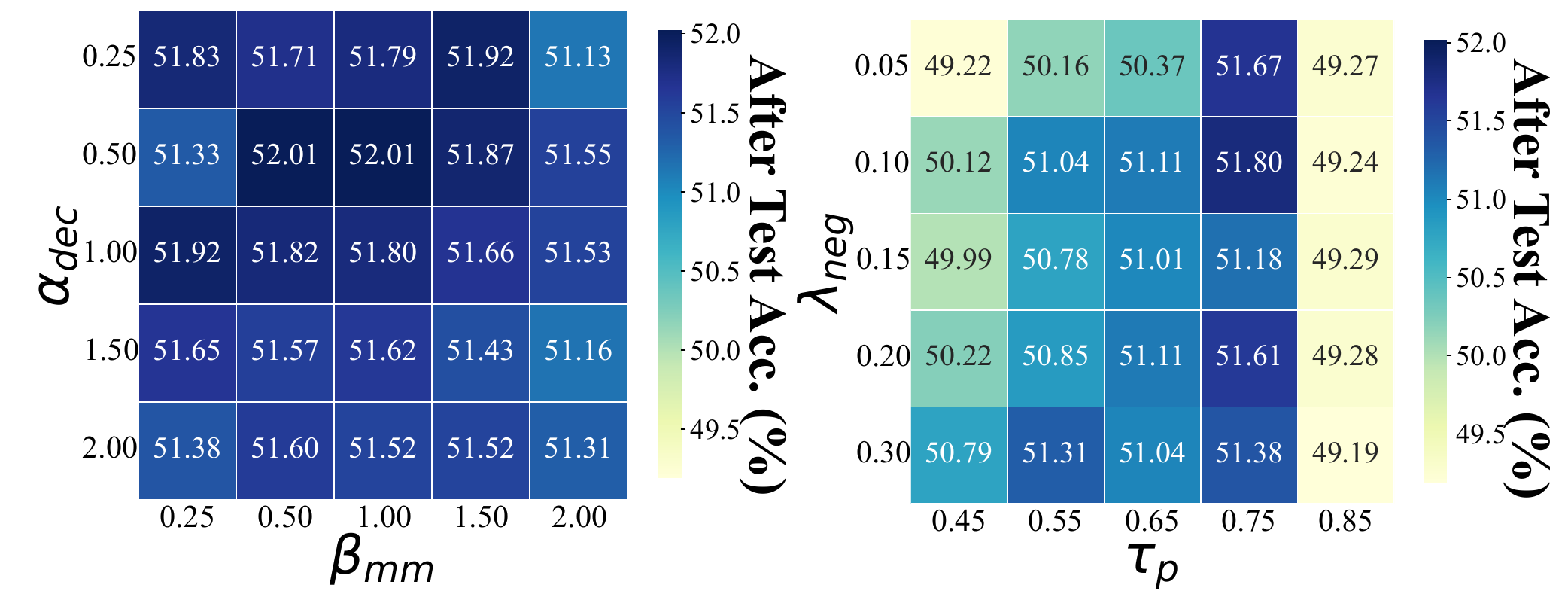}
    \vspace{-17pt}
    \caption{Hyperparameter sensitivity on Movies. Left: sensitivity to $\alpha_{\mathrm{dec}}$ and $\beta_{\mathrm{mm}}$. Right: sensitivity to $\lambda_{\mathrm{neg}}$ and $\tau_p$.}
    \label{fig:mmfgu_hyperparameter}
\end{figure}

\textbf{Hyperparameter Sensitivity.}
Fig.~\ref{fig:mmfgu_hyperparameter} presents a sensitivity analysis of the loss-weight hyperparameters and the probe threshold.
Accuracy remains relatively stable over a wide range of $\alpha_{\mathrm{dec}}$ and $\beta_{\mathrm{mm}}$ values, whereas setting $\tau_p$ to 0.85 leads to consistently inferior performance.
The broad high-accuracy plateau suggests that the method is robust to hyperparameter choices rather than relying on a single finely tuned combination.

\textbf{Plug-in Generality.}
To evaluate whether the proposed design can consistently enhance existing unlearning methods, we integrate it into five runnable backbones. The resulting variants improve retained accuracy for four of the five backbones on average, while all five achieve positive UR gains. These results demonstrate that our design is broadly compatible with heterogeneous unlearning methods and improves their unlearning--utility trade-off. Detailed settings and results are provided in Section~S3 of the supporting material.

\subsection{Efficiency Analysis (Q4)}
\label{sec:mmfgu_q4}

To answer Q4, we evaluate theoretical and empirical computation and communication costs together with update convergence.

\textbf{Efficiency.}
Table~\ref{tab:mmfgu_efficiency} compares theoretical time and space complexity, empirical wall-clock time, and communication cost.\textsc{MMFGU} achieves the lowest wall-clock time and communication cost among all compared methods, including both retraining-based and lightweight baselines.

\begin{table}[H]
    \centering
    \caption{Theoretical time/space complexity and empirical time/communication comparison.}
    \label{tab:mmfgu_efficiency}
    \resizebox{\columnwidth}{!}{%
    \begin{tabular}{l c c c c}
        \specialrule{1.5pt}{1.5pt}{1.5pt}
        \textbf{Method}
        & \textbf{Time Complexity}
        & \textbf{Space Complexity}
        & \textbf{Prac. Time}
        & \textbf{Comm.} \\
        \midrule
        FedEraser
        & $\mathcal{O}(Kmf+rmf^2)$
        & $\mathcal{O}(rmP)$
        & $26.44$ s
        & $2.09$ GB \\

        FedKD
        & $\mathcal{O}(Kmf+Edf^2)$
        & $\mathcal{O}(P)$
        & $1.20$ s
        & $0.05$ GB \\

        FedOSD
        & $\mathcal{O}(Kmf+Kmf^2)$
        & $\mathcal{O}(mP)$
        & $2.71$ s
        & $0.18$ GB \\

        FUSED
        & $\mathcal{O}(Kmf+KmPs)$
        & $\mathcal{O}(P+Ps)$
        & $7.77$ s
        & $0.02$ GB \\

        MoDe
        & $\mathcal{O}\!\left(Kmf+(K_d+K_g)f^2\right)$
        & $\mathcal{O}(2P)$
        & $28.95$ s
        & $1.33$ GB \\

        ReGEnUnlearn
        & $\mathcal{O}(Sf^2+K_rAf^2+qf^2)$
        & $\mathcal{O}(P+SP)$
        & $4.58$ s
        & $0.25$ GB \\

        Retrain
        & $\mathcal{O}(Kmf+Kmf^2)$
        & $\mathcal{O}(mP)$
        & $32.52$ s
        & $1.42$ GB \\

        \textbf{\textsc{MMFGU}}
        & $\mathcal{O}(Sf^2+AKf^2+qP)$
        & $\mathcal{O}(AP+qP)$
        & $\mathbf{0.78}$ s
        & $\mathbf{0.01}$ GB \\
        \specialrule{1.5pt}{1.5pt}{1.5pt}
    \end{tabular}}
\end{table}

\begin{figure}[H]
    \centering
    \includegraphics[width=\columnwidth]{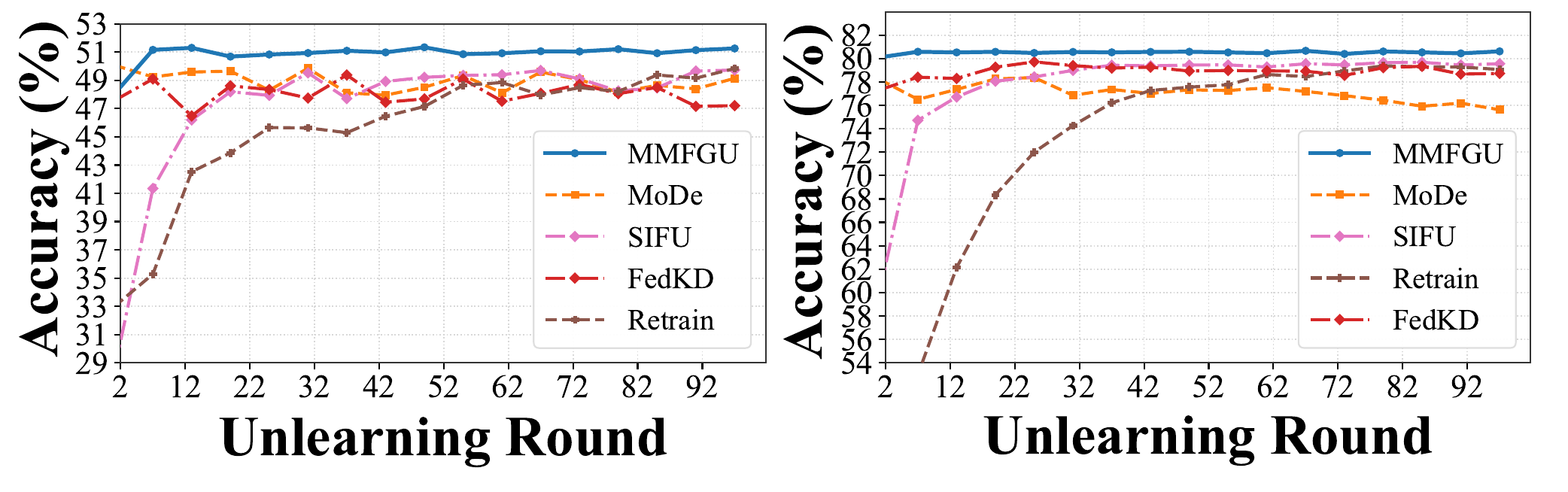}
    \vspace{-20pt}
    \caption{Retained-accuracy convergence over unlearning rounds on Movies (left) and Toys (right).}
    \label{fig:mmfgu_convergence}
\end{figure}

\textbf{Convergence Study.}
Fig.~\ref{fig:mmfgu_convergence} tracks retained accuracy throughout the unlearning process.
On both Movies and Toys, \textsc{MMFGU} reaches a stable high-accuracy region within the first few updates and consistently outperforms the competing methods thereafter.
This complements the endpoint costs in Table~\ref{tab:mmfgu_efficiency} by showing that the efficiency gain is accompanied by stable post-request optimization.

\section{Conclusion}

We presented MMFGU, a multimodal federated graph unlearning framework for Entity/Relation Removal, Modality Removal, and Pairing Removal while preserving retained utility. The key idea is to treat Pairing Removal as an explicit unlearning objective, then combine it with cross-client purge propagation and probe-guided repair. The experimental results show that MMFGU provides a favorable balance among utility, unlearning effectiveness, robustness, and efficiency. Future work can extend the framework to streaming unlearning requests and stronger formal guarantees for multimodal federated graph models.

\bibliography{aaai2026}

\end{document}